\documentclass[times,twocolumn,final,authoryear]{elsarticle}

\usepackage{ycviu}
\usepackage{framed,multirow}

\usepackage{amsmath,amssymb,amsfonts}
\usepackage{graphicx}
\usepackage{textcomp}
\usepackage{tikz}
\usepackage{xcolor}
\usepackage{mathtools}
\usepackage[caption=false]{subfig}
\usepackage{color}
\usepackage{booktabs}
\usepackage{colortbl}
    
\definecolor{ssa_green}{RGB}{137, 209, 42}
\definecolor{tsa_blue}{RGB}{97, 184, 225}
    
\usepackage{graphicx}

\usepackage{pgfplots}
\pgfplotsset{compat=1.12}

\usepackage{amsmath,amssymb} 
\usepackage{xpunctuate}

\usepackage{color, colortbl}
\usepackage{xcolor-material}

\usepackage{soul} 
\usepackage{xspace}
\usepackage{amsmath}

\usepackage{hyphenat}
\DeclareRobustCommand{\new}[1]
{{\textcolor{black}{#1}}}

\usepackage{amssymb}
\usepackage{latexsym}
\usepackage{multirow}
\usepackage{multicol}
\usepackage{url}
\usepackage{xcolor}
\definecolor{newcolor}{rgb}{.8,.349,.1}

\usepackage{soul}

\journal{Computer Vision and Image Understanding}

\begin{document}

%

\ifpreprint
  \setcounter{page}{1}
\else
  \setcounter{page}{1}
\fi

\begin{frontmatter}

\title{Skeleton-based Action Recognition via Spatial and  Temporal Transformer Networks}

\author[1,2]{Chiara \snm{Plizzari}\corref{cor1}} 
\cortext[cor1]{Corresponding author: 
 }
\ead{chiara.plizzari@mail.polimi.it}
\author[1]{Marco \snm{Cannici}}
\author[1]{Matteo \snm{Matteucci}}

\address[1]{Politecnico di Milano, Via Giuseppe Ponzio 34/5, Milan 20133, Italy}
\address[2]{Politecnico di Torino, Corso Duca degli Abruzzi, 24, Turin 10129, Italy}

\begin{abstract}
Skeleton-based Human Activity Recognition has achieved great interest in recent years as skeleton data has demonstrated being robust to illumination changes, body scales, dynamic camera views, and complex background. In particular, Spatial-Temporal Graph Convolutional Networks (ST-GCN) demonstrated to be effective in learning both spatial and temporal dependencies on non-Euclidean data such as skeleton graphs. Nevertheless, an effective encoding of the latent information underlying the 3D skeleton is still an open problem, especially when it comes to extracting effective information from joint motion patterns and their correlations. 
In this work, we propose a novel Spatial-Temporal Transformer network (ST-TR) which models dependencies between joints using the Transformer \textit{self-attention} operator. In our ST-TR model, a Spatial Self-Attention module (SSA) is used to understand intra-frame interactions between different body parts, and a Temporal Self-Attention module (TSA) to model inter-frame correlations.
\new{The two are combined in a two-stream network, whose performance is evaluated on three large-scale datasets, NTU-RGB+D 60, NTU-RGB+D 120, and Kinetics Skeleton 400, consistently improving backbone results. Compared with methods that use the same input data, the proposed ST-TR achieves state-of-the-art performance on all datasets when using joints' coordinates as input, and results on-par with state-of-the-art when adding bones information. }


\end{abstract}


\end{frontmatter}



\section{Introduction}
Human Action Recognition is achieving increasing interest in recent years for the progress achieved in deep learning and computer vision and for the interest of its applications in human-computer interaction, eldercare and healthcare assistance, as well as video surveillance. 
Recent advances in 3D depth cameras such as Microsoft Kinect (\cite{zhang2012microsoft}) and Intel RealSense (\cite{Keselman_2017_CVPR_Workshops}) sensors, and advanced human pose estimation algorithms (\cite{openpose}) made it possible to estimate 3D skeleton coordinates quickly and accurately with cheap devices. 
Nevertheless, several aspects of skeleton-based action recognition still remain open (\cite{a-comprehensive-survey,human-activity,a-survey}). 
The most widespread method to perform skeleton-based action recognition has nowadays become Graph Neural Networks (GNNs), and in particular, Graph Convolutional Networks (GCNs) since, being an efficient representation of non-Euclidean data, they are able to effectively capture spatial (intra-frame) and temporal  (inter-frame) information. Models making use of GCN were first introduced in skeleton-based action recognition by~\cite{yan2018spatial} and they are usually referred to as Spatial-Temporal Graph Convolutional Networks (ST-GCNs). These models process spatial information by operating on skeleton bone-connections along space, and temporal information by considering additional time-connections between each skeleton joint along time. 
Despite being proven to perform very well on skeleton data, ST-GCN models have some structural limitations, some of them already addressed by~\cite{dirgraph,2s-cnn,shift,disent}.

First of all, the topology of the graph representing the human body is fixed for all layers and all the actions; this may prevent the extraction of rich representations for the skeleton movements during time, especially if graph links are directed and information can only flow along a predefined path. Secondly, both Spatial and Temporal Convolution are implemented starting from a standard 2D convolution. As such, they are limited to operate in a local neighborhood, somehow restricted by the convolution kernel size. And finally, as a consequence of the previous, correlations between body joints not linked in the human skeleton, e.g., the left and right hands, are underestimated even if relevant in actions such as ``clapping".

 \begin{figure}[t]
    \centering
    \includegraphics[width=0.46\textwidth]{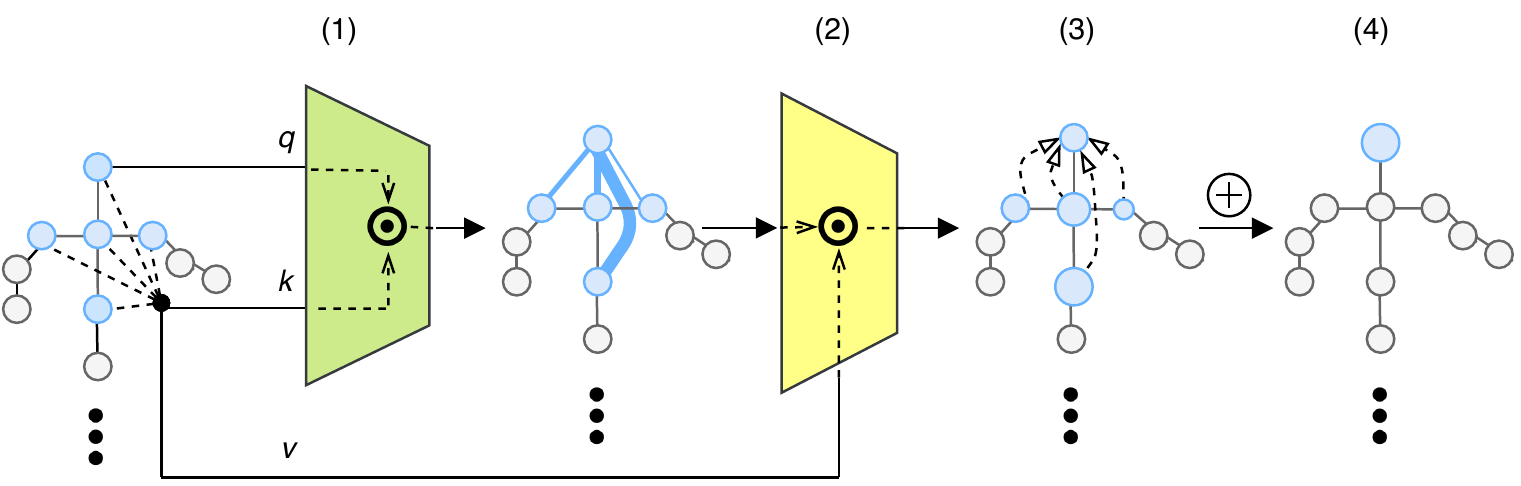}
    \caption{{Self-attention on skeleton joints}. (1) For each body joint, a query $\mathbf{q}$, a key $\mathbf{k}$ and a value vector $\mathbf{v}$ are calculated. (2) Then, the dot product ($\odot$) between the query of the joint and the key of all the other nodes is performed, representing the connection strength between each pair of nodes. (3) Finally, each node is scaled by its correlation w.r.t. the current node, (4) whose new features are obtained summing the weighted nodes together. 
    }
    \label{fig0}
\end{figure}

In this paper, we face all these limitations by employing a modified Transformer self-attention operator, as depicted in Figure \ref{fig0}. Despite being originally designed for Natural Language Processing (NLP) tasks, \new{Transformer self-attention has shown remarkable results on a broad range of computer vision tasks, spanning from classical classification and detection (\cite{dosovitskiy2020image,Bello_2019_ICCV,non-local,carion2020end}), to more complex tasks such as those involving point clouds (\cite{zhao2020point}), generative modeling (\cite{oord2016conditional,parmar2018image}) and captioning (\cite{he2020image}).}
In our setting, the sequentiality and hierarchical structure of human skeleton sequences, as well as the flexibility of Transformer self-attention (\cite{attention}) in modeling long-range dependencies, make the Transformer a perfect solution to tackle ST-GCN weaknesses. 
In our work, we aim to apply Transformer to spatial-temporal skeleton-based architectures, and in particular to joints representing the human skeleton, with the goal of modeling long-range interactions within human actions both in space, through a Spatial Self-Attention (SSA) module, and time, through a Temporal Self-Attention (TSA) module.
Main contributions of this paper are summarized as follows: 
\begin{itemize}
    \item We propose a novel two-stream Transformer-based model for skeleton activity recognition tasks, employing \textit{self-attention} on both the spatial and the temporal dimensions
    \item We design a \textit{Spatial Self-Attention} (SSA) module to dynamically build links between skeleton joints, representing the relationships between human body parts, conditionally on the action and independently from the natural human body structure. On the temporal dimension, we introduce a  \textit{Temporal Self-Attention} (TSA) module to study the dynamics of a joint along time. We made both layers publicly available for experiments replication and further use \footnote{Code at \color{magenta}\url{https://github.com/Chiaraplizz/ST-TR}}
    \item Our model outperforms ST-GCN (\cite{yan2018spatial}) and A-GCN (\cite{Shi2018TwoStreamAG}) \new{consistently improving backbone results on all datasets and achieving state-of-the-art performance when using joint information, and results on-par with state-of-the-art when bones information is used.} %
    
\end{itemize}

\section{Related Works}

\subsection{\textit{Skeleton-based Action Recognition}}{
Most of the early studies in skeleton-based action recognition relied on handcrafted features (\cite{jointly,points,locations}) exploiting relative 3D rotations and translations between joints. Deep learning revolutionized activity recognition by proposing methods capable of increased robustness (\cite{a-comparative}) 
and able to achieve unprecedented performance. Methods that fall into this category rely on different aspects of skeleton data: (1) Recurrent neural network (RNN) based methods (\cite{Rnn,modeling,global-context,hrnn}) leverage on the sequentiality of joint coordinates, treating input skeleton data as time series. (2) Convolutional neural network (CNN) based methods (\cite{p-cnn,2s-cnn,investigation,liu2017enhanced,bo}) leverage spatial information, in a complementary way to RNN-based ones. Indeed, 3D skeleton sequences are mapped into a pseudo-image, representing temporal dynamics and skeleton joints respectively in rows and columns. (3) Graph neural network (GNN) based methods (\cite{yan2018spatial,li2019actional,Shi2018TwoStreamAG,dirgraph,shift,disent}), make use of both spatial and temporal data by exploiting information contained in the natural topological graph structure of the human skeleton. These latter methods have demonstrated to be the most expressive among the three, and among these, the first model capturing the balance between spatial and temporal dependencies has been the Spatio-Temporal Graph Convolutional Network (ST-GCN) (\cite{yan2018spatial}). 

In this work, we used ST-GCN as the baseline model; its functioning is presented in details in Section~\ref{st-gcn}. Specifically, we propose to substitute regular graph convolutions on both space and time with the transformer self-attention operator. \cite{san} also proposed a Self-Attention Network (SAN) in which embeddings are extracted by segmenting the action sequence in temporal clips, and self-attention is applied among them in order to model long-term semantic information. \new{However, since it applies self-attention on course-grained embeddings rather than skeleton joints, it hardly captures low-level joints' correlations \textit{within} and \textit{between} frames. Instead, by applying self-attention \textit{directly} on nodes in the graph, both intra- and inter-frame, we efficiently model both spatial and temporal dependencies of the skeleton sequence. } }

\subsection{Graph Neural Networks} 
\textit{Geometric deep learning} (\cite{DBLP:journals/corr/BronsteinBLSV16}) refers to all emerging techniques attempting to generalize deep learning models to non-Euclidean domains such as graphs. The notion of Graph Neural Network (GNN) was initially outlined by~\cite{Gori2005ANM} and further elaborated by~\cite{Scarselli2009TheGN}. The intuitive idea underlying GNNs is that nodes in a graph represent objects or concepts while edges represent their relationships. Due to the success of Convolutional Neural Networks, the concept of convolution has later been generalized from grid to graph data. GNNs iteratively process the graph, each time representing nodes as the result of applying a transformation to nodes' and their neighbors' features. The first formulation of CNNs on graphs is due to~\cite{ae482107de73461787258f805cf8f4ed}, who generalized convolution to signals using a~\textit{spectral} construction. This approach had computational drawbacks that have been subsequently addressed by~\cite{DBLP:journals/corr/HenaffBL15} and~\cite{DBLP:journals/corr/DefferrardBV16}. The latter has been further simpliﬁed and extended by \cite{Kipf:2016tc}. A complementary approach is the \textit{spatial} one, where graph convolution is defined as information aggregation (\cite{article7,DBLP:journals/corr/NiepertAK16,DBLP:journals/corr/SuchSDPZMCP17}). In this work we make use of the spectral construction proposed by \cite{Kipf:2016tc}, whose formulation is provided in Section~\ref{st-gcn}.
\begin{figure*}[t]
\centering
\subfloat[Spatial Self-Attention \label{2a}]{
    \begin{minipage}[b]{.39\linewidth}
    \includegraphics[width=\textwidth]{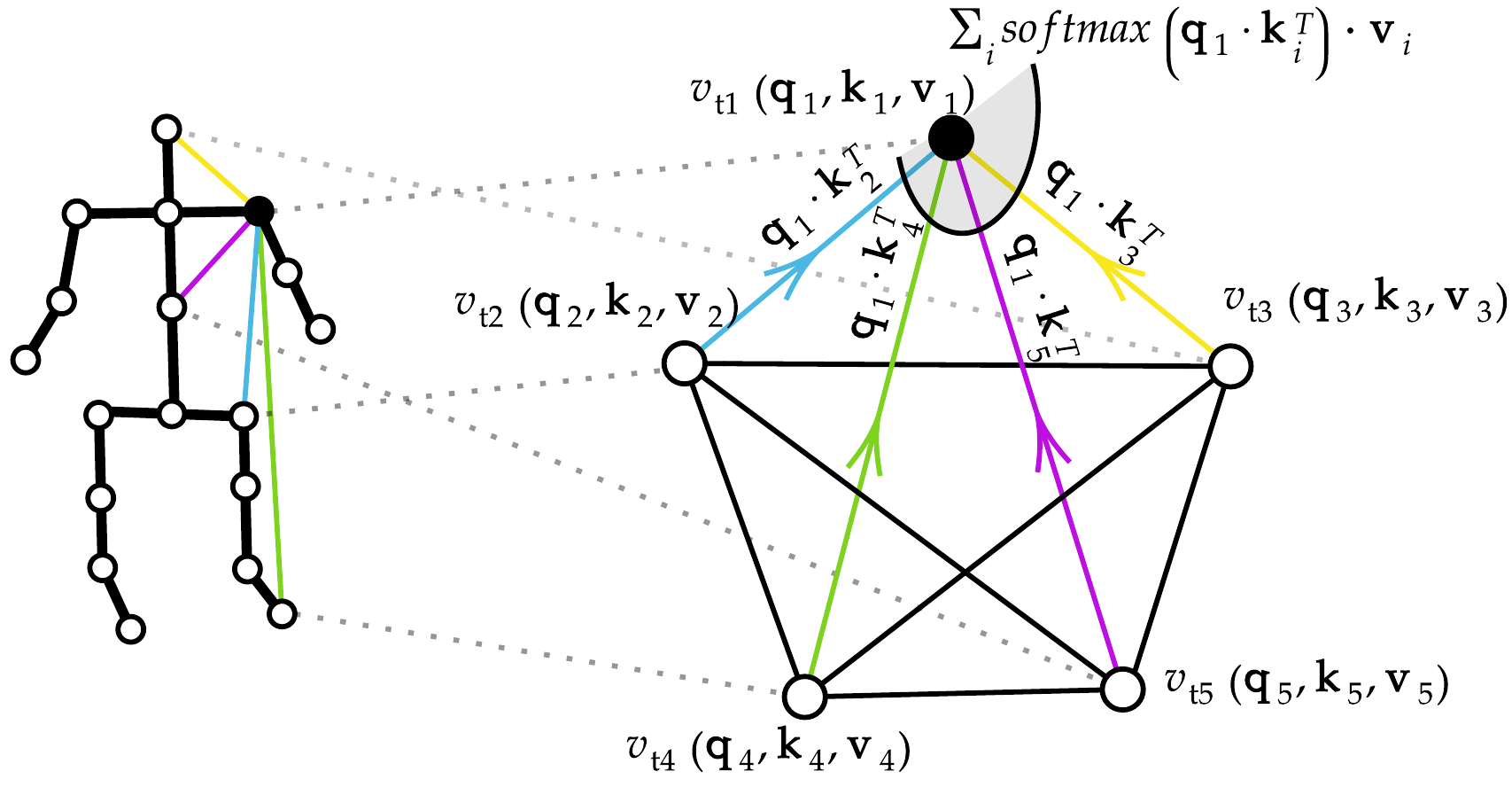}
\end{minipage}}
\hspace{1cm}
\subfloat[Temporal Self-Attention \label{2b}]{\begin{minipage}[b]{.39\linewidth}
    \includegraphics[width=\textwidth]{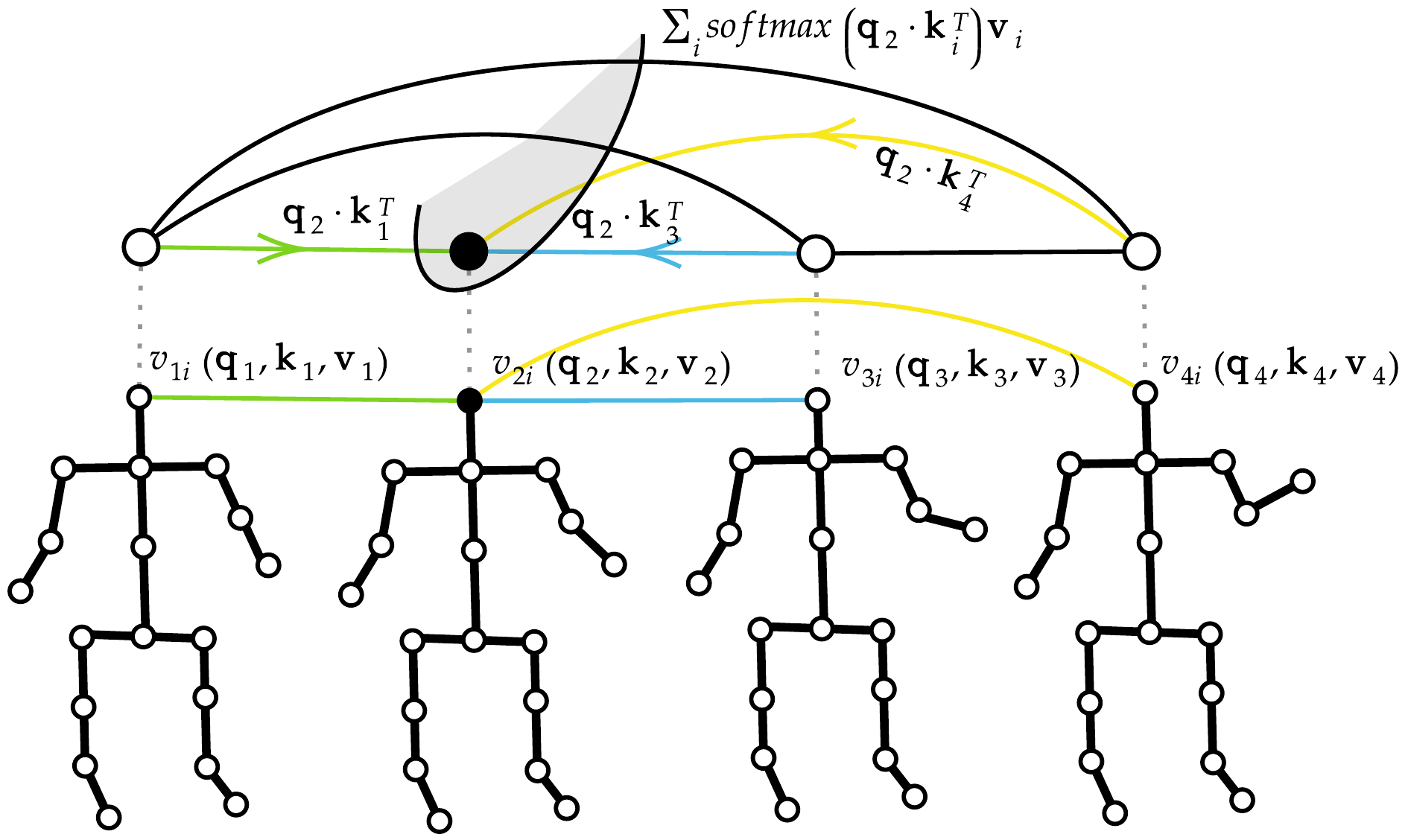}
\end{minipage}}
\caption{{Spatial Self-Attention (SSA)} and {Temporal Self-Attention (TSA)}. Self-attention operates on each pair of nodes, by computing a weight for each of them which represents the strength of their correlation. Those weights are then used to score the contribution of each body joint $v_{ti}$, proportionally to how relevant the node is w.r.t. to all the others. Please notice that on SSA (a), the procedure is illustrated only of a group of five nodes for simplicity, while in practice it operates on all the nodes.}
\label{SSA_TSA}
\end{figure*}

\subsection{\new{Transformers in Computer Vision}}\label{transformer}
The Transformer is the leading neural model for Natural Language Processing (NLP), proposed by \cite{attention} as an alternative to recurrent networks. It has been designed to face two key problems: (i) the processing of very long sequences, which are often intractable both for LSTMs and RNNs, and (ii) the limitations in parallelizing sentence processing, which is usually performed sequentially, word by word, in standard RNNs architectures. The Transformer follows a usual encoder-decoder structure, but it relies solely on \textit{multi-head self-attention} (\cite{attention}). 
\new{Recently, Transformer self-attention has been applied in many popular computer vision tasks. \cite{non-local} proposed a differentiable non-local operator based on self-attention, which allows to capture long-range dependencies both in space and time for a more accurate video classification. After the first attempt of \cite{Bello_2019_ICCV} to use self-attention as an alternative to convolutional operators, \cite{dosovitskiy2020image} proposed a Vision Transformer (ViT), which shows how Transformers can effectively replace standard convolutions on images.  \cite{he2020image} proposed a novel image transformer architecture for the image captioning task.  \cite{carion2020end} made the first attempt to use a Transformer model to tackle detection problems, namely the Detection Transformer (DeTR). 
\cite{zhao2020point} proposed Point Transformer, a model which uses transformer self-attention to encode relations between point clouds, exploiting their permutation invariant nature. Other applications of Transformers in segmentation tasks (\cite{huang2019ccnet}), multi-modal tasks (\cite{lee2020parameter}) and generative modeling (\cite{oord2016conditional,parmar2018image}) have been recently developed, showing the potential of Transformer models on a broad range of tasks.}

\section{Background}
In this section, Spatial-Temporal Graph Convolutional Networks (ST-GCN) by \cite{yan2018spatial} and the original Transformer self-attention by \cite{attention} are summarized, being the basic blocks of the model we propose in this paper.

\subsection{Skeleton Sequences Representation}

Given a sequence of skeletons, we define $V$ as the number of joints representing each skeleton and $T$ as the total number of skeletons composing the sequence, also named frames in the following.
In order to represent the sequence, a spatial temporal graph is built, i.e., $G=(N,E)$, where $N=\{v_{ti}|t=1,...,T,i=1,...,V\}$ represents the set of all the nodes $v_{ti}$ of the graph, i.e., the body joints of the skeleton along all the time sequence, and $E$ represents the set of all the connections between nodes. $E$ consists of two subsets; the first subset $E_S=\{(v_{ti},v_{tj}) \mid i,j = 1,\dots,V, t=1,\dots,T\}$ is composed by the intra-skeleton connections at each time interval $t$, for any pair of joints $(i,j)$ connected by a bone in the human skeleton. The subset $E_S$ of intra-skeleton connections is commonly further divided into $K$ disjoint partitions, based on some criterion (\cite{yan2018spatial}) (e.g., distance from the center of gravity), and encoded using a set of adjacency matrices $\tilde{\mathbf{A}}_k \in \{0,1\}^{V \times V}$. The second subset $E_T=\{(v_{ti},v_{(t+1)i}) \mid i=1,\dots,V, t=1,\dots,T\}$ consists of all the inter-frame connections between joints along consecutive time frames. The result is a graph extending on both the spatial and the temporal dimension.

\subsection{Spatial Temporal Graph Convolutional Networks}\label{st-gcn}

Spatial Temporal Graph Convolutional Networks (ST-GCN) have been introduced by~\cite{yan2018spatial}. A ST-GCN is structured as a hierarchy of stacked spatial-temporal blocks, which are internally composed of a spatial convolution (GCN) followed by a temporal convolution (TCN). 

The spatial sub-module uses the Graph Convolution formulation proposed by~\cite{Kipf:2016tc}, which can be summarized as it follows:
\begin{gather}
\label{eq1}
\textbf{f}_{out}=\sum^{K_s}_k(\textbf{f}_{in}\mathbf{A}_k)\textbf{W}_k, \\
\mathbf{A}_k=\textbf{D}_k^{-\frac{1}{2}} ({{{{\tilde{\textbf{A}}_k}+\textbf{I}}}}) \textbf{D}_k^{-\frac{1}{2}}, D\textsubscript{ii}=\sum^{K_s}_k (\tilde{\mathbf{A}}^{ij}_k+\mathbf{I}\textsubscript{ij}),
\end{gather}
where $K_s$ is the kernel size on the spatial dimension, ${\tilde{\textbf{{A}}}_k}$ is the adjacency matrix of the undirected graph representing intra-body connections, \textbf{I} is the identity matrix and $\textbf{W}_k$ is a trainable weight matrix. The temporal convolution sub-module (TCN) is implemented as a $1 \times K_t$ 2D convolution operating on $(V, T)$ dimensions of the $(C_{in}, V, T)$ input volume, where $K_t$ is the number of frames considered within the kernel receptive field. 

As shown in Equation \ref{eq1}, the graph structure is predefined, being the adjacency matrix fixed. In order to make it \textit{adapative}, \cite{Shi2018TwoStreamAG} introduced the \textit{Adaptive Graph Convolutional Network} (A-GCN), where the GCN formulation in Equation \ref{eq1} is replaced by the following:
\begin{equation}
    \textbf{f}_{out}=\sum^{K_s}_k\textbf{f}_{in}(\mathbf{A}_k+\mathbf{B}_k+\mathbf{C}_k)\textbf{W}_k ,
\end{equation}
where ${\textbf{A}}_k$ is the same as the one in Equation \ref{eq1}, $\textbf{B}_k$ is learned during training, and $\textbf{C}_k$ determines whether two vertices are connected or not through a similarity function. 

\subsection{Transformer Self-Attention}
The original Transformer model of~\cite{attention} employs \textit{self-attention}, i.e., a \textit{non-local operator} originally designed to operate on words in NLP tasks with the goal of enriching the embedding of each word based on the surrounding context. In the Transformer, new word embeddings are computed by comparing pairs of words an then mixing their embeddings together based on how much a word is relevant w.r.t. the others. By gathering clues from the surrounding context, self-attention enables to extract a better meaning from each word, dynamically building relations within and between phrases.

In particular, for each word embedding $\mathbf{w}\textsubscript{i} \in W=\{\mathbf{w}_1,..., \mathbf{w}_n\}$, a query $\mathbf{q} \in \mathbb{R}^{d_q}$, a key $\mathbf{k} \in \mathbb{R}^{d_k}$ and a value vector $\mathbf{v} \in \mathbb{R}^{d_v}$ are computed through trainable linear transformations, 
independently. Then, a score for each word embedding is obtained by taking the dot product $\alpha_{ij} = \mathbf{q}_i \cdot {\mathbf{k}}^T_{j}$ $\forall i,j=1,...,n$, where $n$ is the total number of nodes being considered. This score represents how much the word $j$ is relevant for word $i$. To compute the final embedding for word $i$, a weighted sum is computed by first multiplying the value vector of each other word $\mathbf{v}_j$ by the corresponding score $\alpha_{ij}$, scaled through the softmax function, and then summing these vectors together. This process, also called \textit{scaled dot-product attention}, can be written in matrix form as it follows:
%
\begin{equation}
Attention(\mathbf{Q}, \mathbf{K}, \mathbf{V})=softmax \left(\frac{\mathbf{QK}\textsuperscript{T}}{\sqrt{d\textsubscript{k}}} \right)\mathbf{V},
\end{equation}
where $\mathbf{Q}$, $\mathbf{K}$, and $\mathbf{V}$ are matrices containing the predicted query, key and value vectors, respectively, packed together and $d\textsubscript{k}$ is the channel dimension of the key vectors. The division by $\sqrt{d\textsubscript{k}}$ is performed in order to increase gradients stability during training. In order to obtain better performance, a mechanism called \textit{multi-headed attention} is usually applied, which consists in applying attention, i.e., a head, multiple times with different learnable parameters and then finally combining the results.


\section{Spatial Temporal Transformer Network}

We propose the \textit{Spatial Temporal Transformer (ST-TR)} network, an architecture which uses Transformer self-attention to operate on both space and time. We propose to achieve this goal using two modules, the \textit{Spatial Self-Attention (SSA)} and the \textit{Temporal Self-Attention (TSA)} modules, each one focusing on extracting correlations on one of the two dimensions. 

\subsection{Motivation}
The idea behind the original Transformer self-attention is to allow the encoding of both short- and long-range correlations between words in the sentence. Our intuition is that the same approach can be applied to skeleton-based action recognition as well, as correlations between nodes are crucial both on the spatial and on the temporal dimension. 
We consider the joints comprising the skeleton as a bag-of-words and make use of the Transformer self-attention to extract node embeddings encoding the relation between surrounding joints, just like words in a phrase in NLP. Contrary to a standard graph convolution, where only the adjacent nodes are compared, we discard any predefined skeleton structure and instead let the Transformer self-attention automatically discover joint relations which are relevant for predicting the current action. The resulting operation acts similarly to a graph convolution, but in which the kernel values are \textit{dynamically predicted} based on the discovered joint relations. The same idea is also applied at the sequence level, by analyzing how each joint changes during the action and building \textit{long-range relations} that span different frames, similarly to how relations between phrases are built in NLP. The resulting operator is capable of obtaining a dynamical representation extending both on the spatial and the temporal dimension. 



 \begin{figure}
    \centering
    \includegraphics[width=0.47\textwidth]{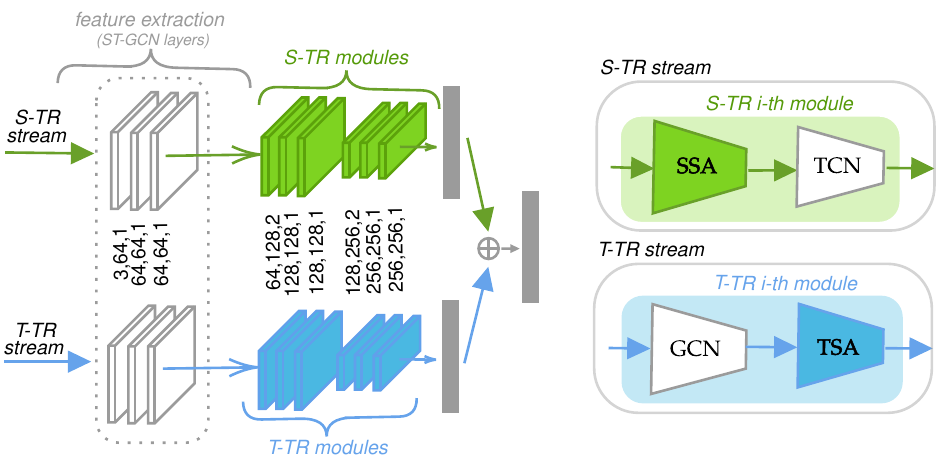}
    \caption{Illustration of two 2s-ST-TR architecture. On each stream, the first three layers extract low level features through standard ST-GCN (\cite{yan2018spatial}) layers. At each successive layer, on the S-TR stream (coloured in \textcolor{ssa_green}{green}), SSA is used to extract spatial information, followed by a 2D convolution on time dimension (TCN), while on the T-TR stream  (coloured in \textcolor{tsa_blue}{blue}), TSA is used to extract temporal information, while spatial features are extracted by a standard graph convolution (GCN).}
    \label{architecture}
\end{figure}

\subsection{Spatial Self-Attention (SSA)}
\label{sec:ssa-descr}

The Spatial Self-Attention module applies self-attention \textit{inside each frame} to extract low-level features embedding the relations between body parts. This is achieved by computing correlations between each pair of joints in every single frame independently, as depicted in Figure~\ref{SSA_TSA}a. 
Given the frame at time $t$, for each node $v_{ti}$ of the skeleton, a \textit{query} vector  $\mathbf{q}^t_i \in \mathbb{R}^{dq}$, a \textit{key} vector $\mathbf{k}^t_i \in \mathbb{R}^{dk}$ and a \textit{value} vector $\mathbf{v}^t_i \in \mathbb{R}^{dv}$ are first computed by applying trainable linear transformations to the node features $\mathbf{n}_i^t \in \mathbb{R}^{C_{in}}$ with parameters $\mathbf{W}_q \in \mathbb{R}^{C_{in} \times dq}$, $\mathbf{W}_k \in \mathbb{R}^{C_{in} \times dk}$, $\mathbf{W}_v \in \mathbb{R}^{C_{in} \times dv}$, shared across all nodes. Then, for each pair of body nodes $(v_{ti}, v_{tj})$, a \textit{query-key dot product} is applied to obtain a weight $\alpha^t_{ij}={\mathbf{q}}^t_{i} \cdot {{\mathbf{k}}^t_{j}}^T \in \mathbb{R}, \forall t \in T$ representing the correlation strength between the two nodes. The resulting score $\alpha^t_{ij}$ is used to weight each joint value $\textbf{v}^t_j$, and a weighted sum is computed to obtain a new embedding $\mathbf{z}^t_i$ for node $v_{ti}$, as in the following:
\begin{equation}\label{eq5}
%
    {\mathbf{z}}^t_i=\sum_j{softmax_j \left(\frac{\alpha^t_{ij}}{\sqrt{d\textsubscript{k}}} \right)  {\mathbf{v}}^t_j} ,
\end{equation}
where ${\mathbf{z}^t_i} \in \mathbb{R}^{C_{out}}$ (with $C_{out}$ the number of output channels) constitutes the new embedding of node $v_{ti}$.

Multi-head attention is applied by repeating this embedding extraction process $N_h$ times, each time with a different set of learnable parameters. The set $(\mathbf{z}^t_{i_1}, ..., \mathbf{z}^t_{i_H})$ of node embeddings thus obtained, all referring to the same node $v_{ti}$, is then combined with a learnable transformation, i.e., $concat(\mathbf{z}^t_{i_1},...,\mathbf{z}^t_{i_H})\cdot\mathbf{W}_o$, and constitutes the output features of SSA.

As shown in Figure~\ref{SSA_TSA}a, the relations between nodes (i.e., the $\alpha_{ij}^t$ scores) are dynamically \textit{predicted} in SSA; the correlation structure in the skeleton is then not fixed for all the actions, but it changes adaptively for each sample. SSA operates similar to a graph convolution on a fully connected graph where, however, the kernel values (i.e., the $\alpha_{ij}^t$ scores) are predicted dynamically based on the skeleton pose.

\subsection{Temporal Self-Attention (TSA)}\label{sec:st}

With the Temporal Self-Attention (TSA) module, the dynamics of each joint is studied separately \textit{along all the frames}, i.e., each single joint is considered as independent and correlations between frames are computed by comparing the change in the embeddings of the same body joint along the temporal dimension (see Figure~\ref{SSA_TSA}b). The formulation is symmetrical to the one reported in Equation \eqref{eq5} for SSA:
\begin{equation}
    \alpha^v_{tu} =\mathbf{q}^v_{t} \cdot {\mathbf{k}^v_{u}} \quad \forall v \in V ,\quad
    \mathbf{z}^v_{t} =\sum_j{softmax_u\left(\frac{\alpha^v_{tu}}{\sqrt{d\textsubscript{k}}}\right)  \mathbf{v}^v_u},
\end{equation}
where $v_{ti}, v_{ui}$ indicate the same joint $v$ in two different instants $t, u$, $\alpha^i_{tu} \in \mathbb{R}$ is the correlation score, $\mathbf{q}^i_t \in \mathbb{R}^{dq}$ is the query associated to $v_{ti}$, $\mathbf{k}^i_u \in \mathbb{R}^{dk}$ and $\mathbf{v}^i_u \in \mathbb{R}^{dv}$ are the key and value associated to joint $v_{ui}$ (all computed using trainable linear transformations as in SSA), and $\mathbf{z}^i_t \in \mathbb{R}^{C_{out}}$ is the resulting node embedding. Note that  the notation used in this section is opposite w.r.t. the one used in Section \ref{sec:ssa-descr}; subscripts indicate time while superscripts indicate the joint. Multi-head attention is applied in TSA as in SSA. An example of TSA is depicted in Figure~\ref{SSA_TSA}b.

The TSA module, by extracting inter-frame relations between nodes in time, can learn how to correlate frames apart from each other (e.g., nodes in the first frame with those in the last one), capturing discriminant features that are not otherwise possible to capture with a standard ST-GCN convolution, being this limited by the kernel size.

\begin{figure}[t]
    \centering
    \includegraphics[width=0.37\textwidth]{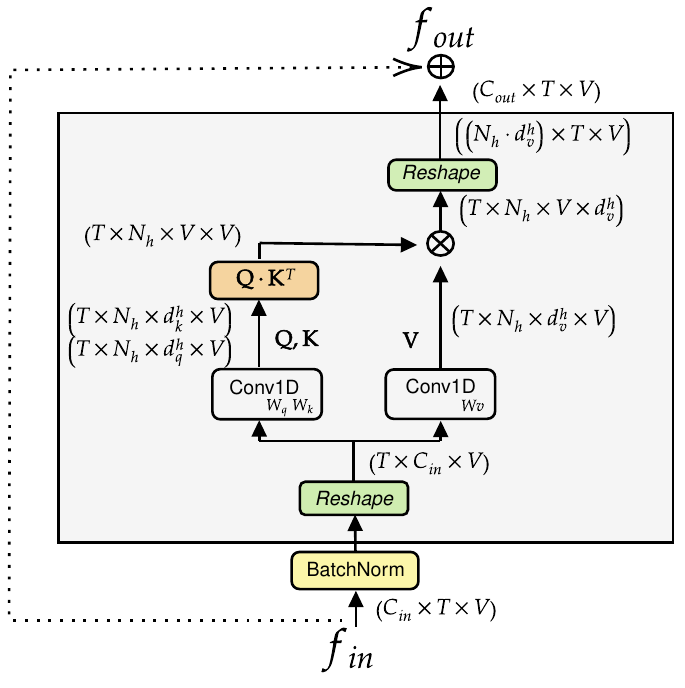}
    \caption{Illustration of a SSA module (the implementation of TSA is the same, with the only difference that the dimension $V$ corresponds to $T$ and viceversa). The input $f_{in}$ is reshaped by moving $T$ in the batch dimension, such that self-attention operates on each time frame separately. SSA is implemented as a matrix multiplication, where $\mathbf{Q}, \mathbf{K}$ and $\mathbf{V}$ are the query, key and value matrix respectively, and $\otimes$ denotes the matrix multiplication. }
    \label{implementation}
\end{figure}

\subsection{Two-Stream Spatial Temporal Transformer Network}\label{2s}

To combine the SSA and TSA modules, a two-stream architecture named ST-TR is used, as similarly proposed by~\cite{Shi2018TwoStreamAG} and~\cite{dirgraph}. In our formulation, the two streams differentiate on the way the proposed self-attention mechanisms are applied: SSA operates on the spatial stream (named S-TR), while TSA on the temporal one (named T-TR). 
On both streams, 
node features are first extracted by a three-layers residual network, where each layer processes the input on the spatial dimension through graph convolution (GCN), and on the temporal dimension through a standard 2D convolution (TCN), as done by~\cite{yan2018spatial}\footnote{In principle other features, e.g., visual features, could be added here but we want in this paper to focus on pure skeleton base action recognition and we leave this option for future investigations.}. 
SSA and TSA are then applied on the S-TR and on the T-TR streams in the subsequent layers in substitution to the GCN and TCN feature extraction modules respectively (Figure~\ref{architecture}). The S-TR stream and T-TR stream are end-to-end trained separately along with their corresponding feature extraction layers. The sub-networks outputs are eventually fused together by summing up their softmax output scores to obtain the final prediction, as proposed by \cite{Shi2018TwoStreamAG} and \cite{dirgraph}.

\paragraph*{\textbf{Spatial Transformer Stream (S-TR)}} In the spatial stream, self-attention is applied at the skeleton level through a SSA module, which focuses on spatial relations between joints. The output of the SSA module is passed to a 2D convolutional module with kernel $K_t$ on the temporal dimension (TCN), as done by \cite{yan2018spatial}, in order to extract temporally relevant features, as shown in Figure \ref{architecture} and expressed in the following: 
\begin{equation}
    \textbf{S-TR}(x)= Conv_{2D(1\times{K_t})}(\textbf{{SSA}}(x)) .
\end{equation}
Following the original Transformer, the input is passes 
through a Batch Normalization layer~(\cite{ioffe2015batch, bn_tr}), and skip connections are used to sum the input to the output of the SSA module (see Figure \ref{implementation}).

\paragraph*{\textbf{Temporal Transformer Stream (T-TR)}} 
The temporal stream, instead, focuses on discovering inter-frame temporal relations. Similarly to the S-TR stream, inside each T-TR layer, a standard graph convolution sub-module (\cite{yan2018spatial}) is followed by the proposed Temporal Self-Attention module:
\begin{equation}
    {\textbf{T-TR}}(x)= \textbf{TSA}(GCN(x)) .
\end{equation}
TSA operates on graphs linking the same joint along all the time dimension (e.g., all left feet, or all right hands).

\subsection{Implementation of SSA and TSA} \label{sec:implementation}

The matrix implementation of SSA (and of TSA) is based on the implementation of Transformer on pixels by~\cite{DBLP:journals/corr/abs-1904-09925}. As shown in Figure~\ref{implementation}, given an input tensor of shape $(C_{in},T,V)$, where $C_{in}$ is the number of input features, $T$ is the number of frames and $V$ is the number of nodes, a matrix $\mathbf{X}_V \in \mathbb{R}^{T \times C_{in} \times V}$ is obtained by rearranging the input. Here the $T$ dimension is moved inside the batch dimension, effectively implementing parameter sharing along the temporal dimension and applying the transformation separately on each frame:
\begin{equation}\label{tr_con}
\begin{aligned}
&head_h(\mathbf{X}_V)= Softmax \left( \frac{(\mathbf{X}_V\mathbf{W}\textsubscript{q})(\mathbf{X}_V\mathbf{W}\textsubscript{k})\textsuperscript{T}}{\sqrt{d^h_{k}}} \right)(\mathbf{X}_V\mathbf{W}_v) \\
&SelfAttention_V=Concat(head\textsubscript{1},...,head_{N_h})\mathbf{W}\textsuperscript{o}  ,
\end{aligned}
\end{equation}
%
where the product with $\mathbf{W}_q \in \mathbb{R}^{C_{in} \times N_h \times d^h_q}$, $\mathbf{W}_k \in \mathbb{R}^{C_{in} \times N_h \times d^h_k}$ and $\mathbf{W}_v \in \mathbb{R}^{C_{in} \times N_h \times d^h_v}$ gives rise respectively to $\mathbf{Q} \in \mathbb{R}^{T \times N_h \times d^h_q \times V}$, $\mathbf{K} \in \mathbb{R}^{T \times N_h \times d^h_k \times V}$ and $\mathbf{V} \in \mathbb{R}^{T \times N_h \times d^h_v \times V}$, being $N_h$ the number of heads, and $\mathbf{W}^o$ a learnable linear transformation combining the heads outputs. 
The output of the Spatial Transformer is then rearranged back into $\mathbb{R}^{C_{out} \times T \times V}$. The TSA matrix implementation has the same expression as Equation \eqref{tr_con}, differing only in the way the input $\mathbf{X}$ is processed. Indeed, in order to be processed by each TSA module, the input is reshaped into a matrix $\mathbf{X}_T \in \mathbb{R}^{V \times C_{in} \times T}$, where the $V$ dimension has been moved in the first position and aggregated to the batch dimension, not reported here explicitly, in order to operate separately on each joint along the time dimension. 
The formulation is analogous to Equation \eqref{tr_con}, differing only in the shape of matrices, which become $\mathbf{Q} \in \mathbb{R}^{V \times N_h \times d^h_q \times T}$, $\mathbf{K} \in \mathbb{R}^{V \times N_h \times d^h_k \times T}$ and $\mathbf{V} \in \mathbb{R}^{V \times N_h \times d^h_v \times T}$.


\section{Model Evaluation}
To understand the impact of both the Spatial and Temporal Transformer streams, we analyze their performance separately and in different configurations through extensive experiments on NTU-RGB+D 60 (\cite{ntu}) (see Table \ref{table:1a}-\ref{table:ntu60}). Then, for a comparison with the state-of-the-art, we test the resulting best configurations on the Kinetics dataset (\cite{Kin}) 
and on the NTU-RGB+D 120 dataset (\cite{ntu120}), which represents to date one of the most complex skeleton-based action recognition benchmarks (see Table \ref{table:ntu120}-\ref{table:kinetics}).

\subsection{Datasets}

\paragraph*{\textbf{\textit{NTU RGB+D 60 and NTU RGB+D 120}}}{The NTU RGB+D 60 (NTU-60) dataset is a large-scale benchmark for 3D human action recognition collected using Microsoft Kinect v2 by \cite{ntu}. 
Skeleton information consists of 3D coordinates of $25$ body joints and a total of $60$ different action classes. The NTU-60 dataset follows two different criteria for evaluation. The first one, called \textit{Cross-View Evaluation} (X-View), uses $37,920$ training and $18,960$ test samples, split according to the camera views from which the action is taken. The second one, called \textit{Cross-Subject Evaluation} (X-Sub), is composed instead of $40,320$ training and $26,560$ test samples. Data collection has been performed with $40$ different subjects performing actions and divided into two groups, one for training and the other for testing. NTU RGB+D 120 (\cite{ntu120}) (NTU-120) is an extension of NTU-60, which adds $57,367$ new skeleton sequences representing $60$ new actions. 
To perform the evaluation, the extended dataset follows two criteria: the first one is the \textit{Cross-Subject Evaluation (X-Sub)}, the same used for NTU-60, while the second one is called \textit{Cross-Setup Evaluation (X-Set)}, which substitutes Cross-View by splitting training and testing samples based on the parity of the camera setup IDs.  } 

\paragraph*{\textbf{\textit{Kinetics}}} The Kinetics skeleton dataset (\cite{yan2018spatial}) is obtained by extracting skeleton annotations from videos composing the Kinetics $400$ dataset (\cite{Kin}), by using the OpenPose toolbox (\cite{openpose}). It consists of $240,436$ training and $19,796$ testing samples, representing a total of $400$ action classes. Each skeleton is composed by 18 joints, each one provided with the 2D coordinates and a confidence score. For each frame, a maximum of 2 people are selected based on the highest confidence scores. 

\subsection{Model Complexity}\label{sec:complexity}
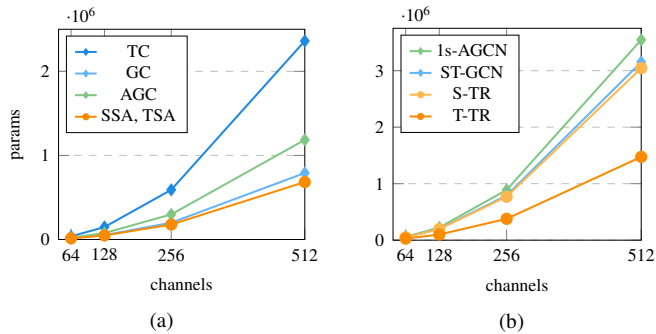
\begin{figure}[t]

\centering
    \begin{minipage}{.49\linewidth}

    \subfloat[]{
    \resizebox {\linewidth} {!} {
        \begin{tikzpicture}

        \begin{axis}[
          legend style={nodes={scale=0.7, transform shape}},
          every axis plot/.append style={thick},
        every mark/.append style={mark size=50pt},
          legend image post style={scale=0.5},
           ticklabel style = {font=\scriptsize},
           label style={font=\scriptsize},
          enlargelimits=false,
          ylabel={params},
          xlabel={channels},
           xmin=40, xmax=512,
           ymin=0, ymax=2500000,
          xtick={64,128, 256, 512},
          legend pos=north west,
          ymajorgrids=true,
          grid style=dashed,
          width=5cm,
          height=4.5cm
        ]
        
      \addplot[
      color=MaterialBlue600,
      mark=diamond*]
    table[x index=0,y index=1,col sep=comma]
    {plots/linea4.txt};
    \addlegendentry{TC} 
    
        \addplot[
          color=MaterialBlue300,
          mark=diamond*]
        table[x index=0,y index=1,col sep=comma]
        {plots/linea1.txt};
        \addlegendentry{GC} 

     \addplot[
          color=MaterialGreen300,
          mark=diamond*]
        table[x index=0,y index=1,col sep=comma]
        {plots/agcn.txt};
        \addlegendentry{AGC}

    \addplot[
      thick,
      color=MaterialOrange600,
      mark=oplus*]
    table[x index=0,y index=1,col sep=comma]
    {plots/linea3.txt};
    \addlegendentry{SSA, TSA}

        \end{axis}

        \end{tikzpicture}%
      }
    \label{fig:ssa_vs_gc_params}
    }
\end{minipage}
\hfill
\begin{minipage}{.45\linewidth}
\subfloat[]{
\resizebox {\columnwidth} {!} {
    \begin{tikzpicture}
    \begin{axis}[
      legend style={nodes={scale=0.7, transform shape}},
      every axis plot/.append style={thick},
        every mark/.append style={mark size=50pt},
      legend image post style={scale=0.5},
      ticklabel style = {font=\scriptsize},
      label style={font=\scriptsize},
      enlargelimits=false,
      xlabel={channels},
       xmin=40, xmax=512,
       ymin=0, ymax=3750000,
      xtick={64,128, 256, 512},
      legend pos=north west,
      ymajorgrids=true,
      grid style=dashed,
      width=5cm,
      height=4.5cm
    ]

 \addplot[
      color=MaterialGreen300,
      mark=diamond*]
    table[x index=0,y index=1,col sep=comma]
    {plots/agcn+conv.txt};
        \addlegendentry{1s-AGCN} 
    
    \addplot[
      color=MaterialBlue300,
      mark=diamond*]
    table[x index=0,y index=1,col sep=comma]
    {plots/conv+conv.txt};
    \addlegendentry{ST-GCN} 
    
    \addplot[
    thick,
      color=MaterialOrange300,
      mark=oplus*]
    table[x index=0,y index=1,col sep=comma]
    {plots/linea6.txt};
        \addlegendentry{S-TR} 

    \addplot[
    thick,
      color=MaterialOrange600,
      mark=oplus*]
    table[x index=0,y index=1,col sep=comma]
    {plots/linea7.txt};
        \addlegendentry{T-TR}

    \end{axis}
    \end{tikzpicture}%
  }
  \label{fig:nets_params}
  }
\end{minipage}
\begin{minipage}[t]{.05\linewidth}\end{minipage}
\caption{(a) Difference in terms of parameters between Graph Convolution (GC), Adaptive Convolution (AGC), Spatial Self-Attention (SSA) modules of $C_{in}=C_{out}$ channels, and between Temporal Convolution (TC) and Temporal Self-Attention (TSA) modules; (b) parameters comparison between ST-GCN, 1s-AGCN and our novel S-TR and T-TR. Best viewed in colors.}
\label{fig:ssa_tsa_params}
\end{figure}

We perform an analysis on the complexity of the different self-attention modules we designed, and compare them to ST-GCN modules (\cite{yan2018spatial}), based on standard convolution, and to 1s-AGCN (\cite{Shi2018TwoStreamAG}) modules, based on adaptive graph convolution. 
First, we compare in Figure~\ref{fig:ssa_tsa_params}a, singularly, a layer of standard convolution with our transformer mechanism, setting $C_{in}=C_{out}$ channels. This results in the same number of parameters for both TSA and SSA, since the convolutions performed internally have the same kernel dimensions and both the query-key dot product and the logit-value product are parameter free.
It can be seen that SSA introduces less parameters than GC, especially when dealing with a large number of channels, where the maximum $\Delta_{GC-SSA}$, i.e., the decrease in terms of parameters, is $1.1 \times 10^5$. When dealing with adaptive modules (AGC), an additional number of parameters has to be considered, resulting in a difference with respect to SSA of $\Delta_{AGC-SSA}=5\times 10^5$. 
On the temporal dimension $\Delta_{TC-TSA}$ reaches a value of $16.8 \times 10^5$. Temporal convolution in~\cite{yan2018spatial} is implemented as a $2D$ convolution with filter $1 \times F$, where $F$ is the number of frames considered along the time dimension, and it is usually set to $9$, striding along $T=300$ frames. Thus, substituting it with a self-attention mechanism results in a great complexity reduction, in addition to better performance, as reported in the next sections. 


Finally, in Figure \ref{fig:nets_params} we also compare the entire stream architectures, i.e., ST-GCN (\cite{yan2018spatial}) and 1s-AGCN (\cite{Shi2018TwoStreamAG}) with the proposed S-TR and T-TR streams in terms of parameters.
As expected from the considerations above, the biggest improvement in parameters reduction is achieved by substituting temporal convolution with TSA, i.e., in T-TR, with a $\Delta_{ST-GCN--T-TR}=16.7\times10^5$. On the spatial dimension the difference in terms of parameters is not as pronounced as in temporal dimension, but it is still significant, with a $\Delta_{ST-GCN--S-TR}=1.07\times10^5$ and $\Delta_{1s-AGCN--S-TR}=5.0\times10^5$.

\subsection{{Experimental Settings}} Using PyTorch (\cite{paszke2019pytorch}) framework, we trained our models for a total of $120$ epochs with batch size $32$ and SGD as optimizer on NTU-60 and NTU-120, while on Kinetics we trained our models for a total of $65$ epochs, with batch size $128$. The learning rate is set to $0.1$ at the beginning and then reduced by a factor of $10$ at the epochs \{$60$, $90$\} and \{$45$, $55$\} for NTU and Kinetics respectively. These schedulings have been selected as they have been shown to provide good results on ST-GCN networks used by~\cite{dirgraph}. \new{When using adaptive AGCN modules, we performed a linear warmup of the learning rate during the first epoch.} Moreover, we preprocessed the data with the same procedure used by~\cite{Shi2018TwoStreamAG} and \cite{dirgraph}. In order to avoid overfitting, we also used \textit{DropAttention}, a particular dropout technique introduced by~\cite{Lin2019DropAttentionAR} for regularizing attention weights in Transformer networks, that consists in randomly dropping columns of the attention logits matrix. In all of these experiments, the \textit{number of heads} for multi-head attention is set to $8$, and $d_q, d_k, d_v$ embedding dimensions to $0.25 \times C_{out}$ in each layer, as done in \cite{DBLP:journals/corr/abs-1904-09925}. We did not perform grid search on these parameters. As far as it concerns the model architecture, each stream is composed by 9 layers, of channel dimension $64, 64, 64, 128, 128, 128, 256, 256$ and $256$. Batch normalization is applied to input coordinates, a global average pooling layer is applied before the \textit{softmax} classifier and each stream is trained using the standard cross-entropy loss.


\setlength{\tabcolsep}{1pt}

\begin{table}[t]
\setlength{\tabcolsep}{1.7pt}

    \caption{Comparison between the baseline and our self-attention modules in terms of both performance (accuracy (\%)) and efficiency (number of parameters) on NTU-60 (X-View)}
    \label{table:1a}

    \centering
    \begin{tabular}{lcccc}
    \hline\noalign{\smallskip}
    \textbf{Method} & GCN & TCN &Params [$\times 10^5$] & Top-1\\

    \hline
    ST-GCN  & \checkmark & \checkmark & 31.0 & 92.7 \\
    ST-GCN-fc  & $\mathbf{A}_{fc}$ & \checkmark & 26.5 & 93.7 \\
    1s-AGCN  & $\mathbf{A}_{k}, \mathbf{B}_{k}, \mathbf{C}_{k}$ & \checkmark &34.7& 93.7\\
    1s-AGCN w/o A  & $\mathbf{B}_{k}, \mathbf{C}_{k}$ & \checkmark & 33.1 & 93.4\\
    \hline
    S-TR & SSA & \checkmark & 30.7 & \textbf{94.0} \\
    T-TR& \checkmark & TSA &  17.6 & 93.6 \\

    \hline
    \end{tabular}
%
%
%
%

\end{table}
\subsection{{Results}}

To verify in a fair way the effectiveness of our SSA and TSA modules, we compare the S-TR and T-TR streams individually against the ST-GCN (\cite{yan2018spatial}) baseline (whose results are reported using our learning rate scheduling) and other models that modify its basic GCN module (see Table~\ref{table:1a}): (i) \textit{ST-GCN (fc)}: we implemented a version of ST-GCN whose adjacency matrix is composed of all ones (referred as $\mathbf{A}_{fc}$), to simulate the fully-connected skeleton structure underlying our SSA module and verify the superiority of self-attention over graph convolution on the spatial dimension; (ii) \textit{1s-AGCN}: Adaptive Graph Convolutional Network (AGCN) (\cite{Shi2018TwoStreamAG}) (see Section \ref{st-gcn}), as it demonstrated in the literature to be more robust than standard ST-GCN, in order to remark the robustness of our SSA module over more recent methods; (iii) \textit{1s-AGCN w/o A}: 1s-AGCN without the static adjacency matrix, to verify the effectiveness of our SSA over graph convolution in a similar setting where all the links between joints are exclusively learnt. 
All these methods use the same implementation of convolution on the temporal dimension (TCN). We make a comparison both in terms of model accuracy and number of parameters.
\setlength{\tabcolsep}{0.5pt}
\begin{table}[t]
\centering
\caption{a) Comparison of S-TR and T-TR streams, and their combination (ST-TR) on NTU-60, w and w/o bones. b) Ablations on different model configurations}
\label{table:1}
 \begin{minipage}{0.53\linewidth}
        \subfloat[]{
\begin{tabular}{lccc}

\hline\noalign{\smallskip}
\textbf{Method} & Bones & X-Sub & X-View\\\noalign{\smallskip}
\hline
\noalign{\smallskip}

S-TR &  &86.4 & 94.0 \\
T-TR  &   &86.0 & 93.6 \\
\new{T-TR-agcn}  &   &\new{86.9} & \new{94.7} \\

\hline
ST-TR  & &88.7 & {95.6} \\
\new{ST-TR-agcn}  &   &\new{\textbf{89.2}} & \new{\textbf{95.8}}\\
\hline 

S-TR   & \checkmark &87.9 & 94.9 \\
T-TR  &\checkmark &87.3 & 94.1\\
T-TR-agcn & \checkmark &\new{88.6} &\new{94.7}\\
\hline
ST-TR  & \checkmark& 89.9 & {96.1}\\
ST-TR-agcn & \checkmark &\new{\textbf{90.3}}& \new{\textbf{96.3}}\\

\hline
\end{tabular}
}
\end{minipage}
\hfill
 \begin{minipage}{0.4\linewidth}
        \subfloat[]{
\begin{tabular}{lc}
\hline\noalign{\smallskip}
\textbf{Ablation}  & X-View\\\noalign{\smallskip}
\hline

S-TR-all-layers  &93.3 \\
            T-TR-all-layers & 91.3 \\
            ST-TR-all-layers & 95.0\\
            \hline \noalign{\smallskip}
            S-TR-augmented   & 94.5 \\
            T-TR-augmented  &  90.2 \\
            ST-TR-augmented & 94.9 \\
            \hline \noalign{\smallskip}ST-TR-1s & 93.3 \\
\hline
\new{ST-TR ($k=0$)} & \new{95.4} \\
\new{ST-TR ($k=1$)} & \new{95.6} \\
\new{ST-TR ($k=2$)} & \new{95.7} \\
\hline
\noalign{\smallskip}

\end{tabular}
}
\end{minipage}
\end{table}
\setlength{\tabcolsep}{1.4pt}

Regarding SSA, the performance of S-TR is superior to all methods mentioned above, demonstrating that self-attention can be used in place of graph convolution, increasing the network performance while also decreasing the number of parameters. 
In fact, as it can be seen from Table \ref{table:1a}, S-TR introduces $0.3\times10^5$ parameters less then ST-GCN and $4\times10^5$ less than 1s-AGCN, with a performance increment w.r.t. all GCN configurations. Similarly, regarding TSA, what emerges from the comparison between T-TR and the ST-GCN baseline adopting standard convolution, is that by using self-attention on the temporal dimension the model is significantly lighter ($13.4\times10^5$ less parameters), and achieves an increment in accuracy of $0.9\%$. 

In Table \ref{table:1} we first analyze the performance of the S-TR stream, T-TR stream and their combination by using input data consisting of joint information only. As it can be seen from Table \ref{table:1}a, on NTU-60 the S-TR stream achieves slightly better performance (+0.4\%) than the T-TR stream, on both X-View and X-Sub. This can be motivated by the fact that SSA in S-TR operates on 25 joints only, while on temporal dimension the number of correlations is proportional to the huge number of frames. Again, as shown in Table \ref{table:1a}a, applying self-attention instead of convolution clearly benefits the model on both spatial and temporal dimensions. The combination of the two streams achieves 88.7\% of accuracy on X-Sub and 95.6\% of accuracy on X-View,  outperforming the baseline ST-GCN and surpassing other two-stream architectures (see Table \ref{table:ntu60}).

As adding the differential of spatial coordinates (bones information) demonstrated to lead to better results in previous works (\cite{dirgraph,2s-cnn}), we also studied our Transformer modules on combined joint and bones information. For each node $\mathbf{v}_1=(\mathbf{x}_1, \mathbf{y}_1, \mathbf{z}_1)$ and $\mathbf{v}_2=(\mathbf{x}_2,\mathbf{y}_2,\mathbf{z}_2)$, the bone connecting the two is calculated as 
    $\mathbf{b}_{\mathbf{v}_1,\mathbf{v}_2}=(\mathbf{x}_2-\mathbf{x}_1,\mathbf{y}_2-\mathbf{y}_1,\mathbf{z}_2-\mathbf{z}_1)$.
Both joint and bone information are concatenated along the channel dimension, and then fed to the network. At each layer, the dimension of the input and output channels are doubled as done by \cite{dirgraph} and \cite{2s-cnn}. Results are shown again in Table~\ref{table:1}a, where all previous configurations improve when bones information is added as input. This highlights the flexibility of our method, which is capable of adapting to different input types and network configurations. 

To further test its flexibility, we also perform experiments in which the GCN module is substituted by the AGCN adaptive module on the temporal stream. As it can be seen from Table \ref{table:1}a, these configurations (\textit{T-TR-agcn}) achieve better results than the one using standard GCN on both X-Sub and X-View.



\setlength{\tabcolsep}{4pt}

\subsection{Effect of Applying Self-Attention since Feature Extraction}
We designed our streams to operate starting from high-level features, rather than directly from coordinates, extracted using a sequence of residual GCN and TCN modules as reported in Section \ref{2s}. This set of experiments validates our design choice. In these experiments SSA (TSA) substitutes GCN (TCN) on the S-TR (T-TR) stream, from the very first layer. The configurations reported in Table \ref{table:1}b (named \textit{S-TR-all-layers}), performs worse than the corresponding ones in Table \ref{table:1}a, while still outperforming the baseline ST-GCN (\cite{dirgraph}) (see Table \ref{table:ntu60}). \new{Indeed, self-attention has demonstrated being more efficient when 
incorporated in later stages of the network (\cite{carion2020end,huang2019ccnet,non-local}).} 
Notice that on T-TR, in order to deal with the great number of frames in the very first layers ($T=300$), we divided them into blocks within which SSA is applied, and then gradually reduce the number of blocks 
($d_{block}=10$ where $C_{out}=64$, $d_{block}=10$ where $C_{out}=128$, and a single block of $d_{block}=T^l$ on layers $l$ with $C_{out}=256$).

\new{The standard protocol used in recent works (\cite{shift,Shi2018TwoStreamAG}) that propose alternative modules for ST-GCN based networks is to keep the original ST-GCN backbone architecture fixed in terms of layers composition. Following these works, we 
kept the three original feature extraction layers for a fair comparison. 
We further conduct some targeted experiments in which we vary their number $k$ (Table \ref{table:1}b). As it can be seen, performance is not sensitive to variations of $k$, confirming the effectiveness of the proposed approach.}


\subsection{Effect of Augmenting Convolution with Self-Attention} 
Motivated by the results in \cite{DBLP:journals/corr/abs-1904-09925}, we studied the effect of applying the proposed Transformer mechanism as an augmentation procedure to the original ST-GCN modules. In this configuration, $0.75 \times C_{out}$ features result from GCN (TCN) and they are concatenated to the remaining $0.25 \times C_{out}$ features from SSA (TSA), a setup that
has proven to be effective in \cite{DBLP:journals/corr/abs-1904-09925}. To compensate the reduction of attention channels, \textit{wide attention} is used, i.e., half of the attention channels are assigned to each head, then recombined together while merging heads. The results are reported in Table \ref{table:1}b (referred as \textit{ST-TR-augmented}). Graph convolution is the one that benefits the most from SSA attention (S-TR-augmented, 94.5\%), to be compared with S-TR's 94\% in Table \ref{table:1}a. Nevertheless, the lower number of output features assigned to self-attention prevent temporal convolution improving on T-TR stream. 

\subsection{Effect of combining SSA and TSA in a single stream}
We tested the efficiency of the model when SSA and TSA are combined in a single stream architecture (see Table \ref{table:1}b, referred as \textit{S-TR-1s}). In this configuration, feature extraction is still performed by the original GCN and TCN modules
, while from the 4th layer on, each layer is composed by SSA followed by TSA, i.e., $\textbf{ST-TR-1s}(x)=\textbf{TSA}(\textbf{SSA}(x))$.

We also tested this configuration on NTU-60, obtaining an accuracy of $93.3\%$, slightly lower than the $95.6\%$ accuracy of 2s-ST-TR (see Table \ref{table:1a}a, ST-TR). However, it should be noted that S-TR-1s presents $17.4 \times 10^5$ parameters, drastically reducing the complexity of the baseline ST-GCN which consists in $31 \times 10^5$ parameters. Moreover, it outperforms the ST-GCN baseline by $0.6\%$ using half of the parameters.

\begin{table}
    \caption{Comparison with state-of-the-art accuracy (\%) on NTU-60. Best for both configurations w/ and w/o bones in \textbf{bold}.}
    \label{table:ntu60}
    \centering
    \begin{tabular}{lccc}
    \hline\noalign{\smallskip}
\multicolumn{4}{c}{\textbf{NTU-60}}\\
\cline{1-4}\noalign{\smallskip}
\textbf{Method} & \textbf{Bones} & X-Sub & X-View\\    \noalign{\smallskip}
    \hline
    \noalign{\smallskip}
    STA-LSTM (\cite{sta-lstm})  & &73.4 & 81.2 \\
    VA-LSTM (\cite{va-lstm})  & &79.4 & 87.6 \\
    AGC-LSTM (\cite{att-aug}) & &89.2 & 95.0 \\
    ST-GCN (\cite{yan2018spatial})  & & 81.5 & 88.3\\
        AS-GCN (\cite{li2019actional})&& 86.8 & 94.2 \\

     1s-AGCN (\cite{Shi2018TwoStreamAG}) & & 86.0 & 93.7 \\
      SAN (\cite{san})  &&87.2 & 92.7 \\
    1s Shift-GCN (\cite{shift})  & &87.8 & 95.1 \\
       
\hline
    ST-TR (Ours) & &{88.7} & {95.6} \\
    \new{ST-TR-agcn (Ours)}  &   &\new{\textbf{89.2}} & \new{\textbf{95.8}} \\
\hline 
        2s-AGCN (\cite{Shi2018TwoStreamAG})  & \checkmark& 88.5 & 95.1 \\
    \textcolor{gray}{DGNN (\cite{dirgraph})} &\textcolor{gray}{\checkmark} &  \textcolor{gray}{89.9} & \textcolor{gray}{96.1} \\
2s Shift-GCN (\cite{shift})  & \checkmark &89.7 & 96.0\\
\textcolor{gray}{4s Shift-GCN (\cite{shift})}  & \textcolor{gray}{\checkmark} &\textcolor{gray}{90.7} & \textcolor{gray}{96.5}\\

    MS-G3D (\cite{disent}) &\checkmark& \textbf{91.5} & {96.2} \\
\hline
    ST-TR (Ours) & \checkmark &{{89.9}} & {{96.1}}
    \\
    ST-TR-agcn (Ours) &\checkmark& \new{90.3}&{\textbf{\new{96.3}}}\\

    \hline
    \end{tabular}
\end{table}

\begin{table}[t!]
\setlength{\tabcolsep}{1pt}
\caption{Comparison with state-of-the-art accuracy (\%) of S-TR, T-TR, and their combination (ST-TR) on NTU-120. Best for both configurations w/ and w/o bones in \textbf{bold}.}
\label{table:ntu120}

    \centering
    \begin{tabular}{lccc}
    \hline\noalign{\smallskip}
    \multicolumn{4}{c}{\textbf{NTU-120}}\\
    \cline{1-4}\noalign{\smallskip}
    {\textbf{Method}} & Bones & X-Sub & X-Set\\
    \noalign{\smallskip}
    \hline
    {ST-LSTM (\cite{st-lstm})} && 55.7 & 57.9 \\
    {GCA-LSTM (\cite{gca})} && 61.2 & 63.3 \\
    {RotClips+MTCNN (\cite{lcr})} &&62.2&61.8\\

    {Pose Evol. Map (\cite{bpe})} && 64.6 & 66.9\\
    1s Shift-GCN (\cite{shift}) && 80.9 & 83.2 \\
    \hline
    \noalign{\smallskip}
    {S-TR} (Ours) && 78.6 & 80.7 \\
    {T-TR} (Ours) &&   78.4 & 80.5 \\
    \new{{T-TR-agcn} (Ours)} && \new{80.1} & \new{82.1}\\

    
    \hline
    {ST-TR} (Ours) & &81.9 & 84.1 \\
    ST-TR-agcn (Ours) &&\textbf{82.7} &\new{\textbf{85.0}} \\
    \hline
        \noalign{\smallskip}

    \new{2s-AGCN (\cite{Shi2018TwoStreamAG})} & \checkmark &\new{82.9} & \new{84.9} \\
    \new{2s Shift-GCN (\cite{shift})} &\checkmark& \new{85.3} & \new{86.6}
     \\
    \textcolor{gray}{4s Shift-GCN (\cite{shift}) } & \textcolor{gray}{\checkmark} &\textcolor{gray}{85.9} & \textcolor{gray}{87.6} \\
    \new{MS-G3D (\cite{disent})} & {\checkmark} & \new{\textbf{86.9}} & \new{\textbf{88.4}}\\
    \hline
    \noalign{\smallskip}
    \new{{S-TR} (Ours)} &\checkmark& \new{81.0} & \new{83.6} \\
    \new{{T-TR} (Ours)} &\checkmark&   \new{80.4} & \new{83.0} \\
    \new{{T-TR-agcn} (Ours)} &\checkmark& \new{82.7} & \new{84.9}\\
    \hline
        \noalign{\smallskip}

    \new{{ST-TR} (Ours)} & \checkmark &\new{84.3} & \new{86.7} \\
    \new{ST-TR-agcn (Ours)} &\checkmark&\new{{85.1}} &\new{{87.1}} \\
        \hline

    \end{tabular}

\end{table}

\begin{table}[t!]
\setlength{\tabcolsep}{1pt}
\caption{Comparison with state-of-the-art accuracy (\%) of S-TR, T-TR, and their combination (ST-TR) on Kinetics. Best for both configurations w/ and w/o bones in \textbf{bold}.}
\label{table:kinetics}
    \centering

    \begin{tabular}{lccc}
    \hline\noalign{\smallskip}
    \multicolumn{4}{c}{\textbf{Kinetics}}\\
    \cline{1-4}\noalign{\smallskip}
    \textbf{Method} & \textbf{Bones} & Top-1 & Top-5\\
    \noalign{\smallskip}
    \hline
        
        \multicolumn{2}{l}{ST-GCN (\cite{yan2018spatial})} & 30.7 & 52.8 \\
                AS-GCN (\cite{li2019actional})&& 34.8 & 56.5 \\

        \multicolumn{2}{l}{SAN (\cite{san})} & 35.1 & 55.7\\
    \hline
    \noalign{\smallskip}
    S-TR (Ours)&  & 32.4 & 55.3\\
    T-TR  (Ours)& & 32.4 & 55.2 \\
        \new{T-TR-agcn  (Ours)}& & \new{34.4} & \new{57.1} \\
\hline
    ST-TR (Ours)&
    & 34.5 & 57.6 \\
        \new{ST-TR-agcn  (Ours)}& & \new{\textbf{36.1}} & \new{\textbf{58.7}} \\
    \hline
    \noalign{\smallskip}
            \multicolumn{1}{l}{2s-AGCN (\cite{Shi2018TwoStreamAG})} &\checkmark& 36.1 & 58.7 \\
        \multicolumn{1}{l}{\textcolor{gray}{DGNN (\cite{dirgraph})}} & \textcolor{gray}{\checkmark}&\textcolor{gray}{36.9} & \textcolor{gray}{59.6} \\
        
        \multicolumn{1}{l}{MS-G3D (\cite{disent})} & \checkmark&{38.0} & \textbf{60.9} \\ \hline
    S-TR (Ours)& \checkmark  & 35.4 & 57.9 \\
    T-TR (Ours) & \checkmark & 33.1 &  55.86 \\

    T-TR-agcn (Ours) & \checkmark&  \new{34.7} & \new{56.4} \\
    \hline
        ST-TR (Ours) & \checkmark&  37.0 & 59.7 \\

    ST-TR-agcn (Ours) & \checkmark& \new{\textbf{38.0}} & \new{60.5}\\
    \hline
    \end{tabular}
\end{table}
\section{Comparison with State-Of-The-Art Results}\label{comparison}
In addition to NTU-60, we compare our methods on NTU-120 and Kinetics. \new{For a fair comparison, we compare the ST-TR configurations on methods trained on the same input data (either with joint information only, or both joint and bones information).}
On NTU-60 (Table \ref{table:ntu60}), the proposed ST-TR, when using joint information only, outperforms all the state-of-the-art models using the same type of information. In particular, it outperforms SAN (\cite{san}), another method employing self-attention in skeleton-based action recognition, by up to $3\%$. 
\new{When using bones information, the proposed transformer based architecture outperforms 2s-AGCN (\cite{Shi2018TwoStreamAG}) on both X-Sub and X-Set and our best configuration making use of the AGCN backbone (ST-TR-agcn) reaches performance on-par with state-of-the-art. We further compare ST-TR against 4s Shift-GCN (\cite{shift}) which, in addition to the joint and bones streams, also comprises two extra streams making use of additional temporal information, and DGNN (\cite{dirgraph}), which also makes use of motion information besides joint and bones. 
We report 4s Shift-GCN and DGNN in Table \ref{table:ntu60}-\ref{table:kinetics} with a different color to highlight the difference in the input. ST-TR-agcn outperforms DGNN on both X-Sub and X-View and crucially, although 4s Shift-GCN uses extra input data and combines two additional streams, ST-TR-agcn still achieves on-par results but with a simpler design. }

On NTU-120 (Table \ref{table:ntu120}), the model only based on joints 
outperforms all state-of-the-art methods that use the same information. \new{When adding bones, both ST-TR and ST-TR-agcn outperform 2s-AGCN by up to $3\%$ on both X-Sub and X-Set. Moreover, ST-TR-agcn's results are on-par with 2s Shift-GCN (\cite{shift}) on X-Sub, while they improve on X-Set. 
Our network has only slightly lower performance than MS-G3D (\cite{disent}), which represents to date a very strong baseline in skeleton-based action recognition. Considering that MS-G3D features a multi-path design with multi-scale graph convolutions, the performance obtained by ST-TR is remarkable given that the latter is based on a simpler backbone.}
Finally, on Kinetics (Table \ref{table:kinetics}), our model using only joints outperforms the ST-GCN baseline by $5\%$ \new{and all previous methods using only joint information}. When bones information is added, 
\new{it outperforms both 2s-AGCN and DGNN, and achieves results on-par with the very recent state-of-the-art method MS-G3D.}  


\section{\new{Qualitative Results}}
\begin{figure}[t]
    \centering
    \includegraphics[width=0.43\textwidth]{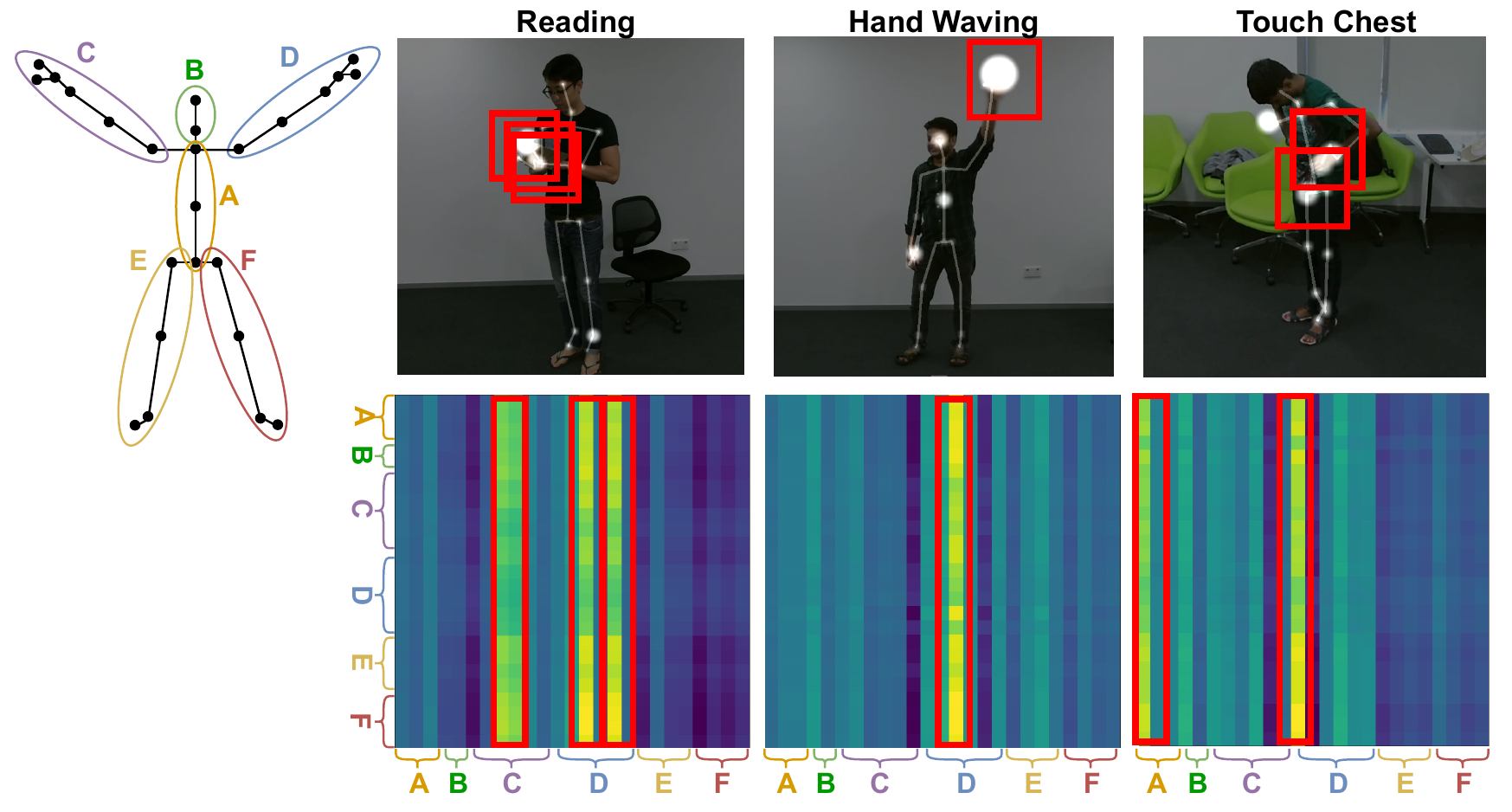}
    \caption{\new{Skeleton of the subjects performing the action (top) and the corresponding SSA heatmaps (bottom). We display with red boxes the joints which the network identifies as the most relevant, while the corresponding spatial self-attention scores are highlighted in the attention maps. } 
    }
    \label{fig:heatmaps}
\end{figure}

\begin{figure}[t]
    \centering
    \includegraphics[width=0.43\textwidth]{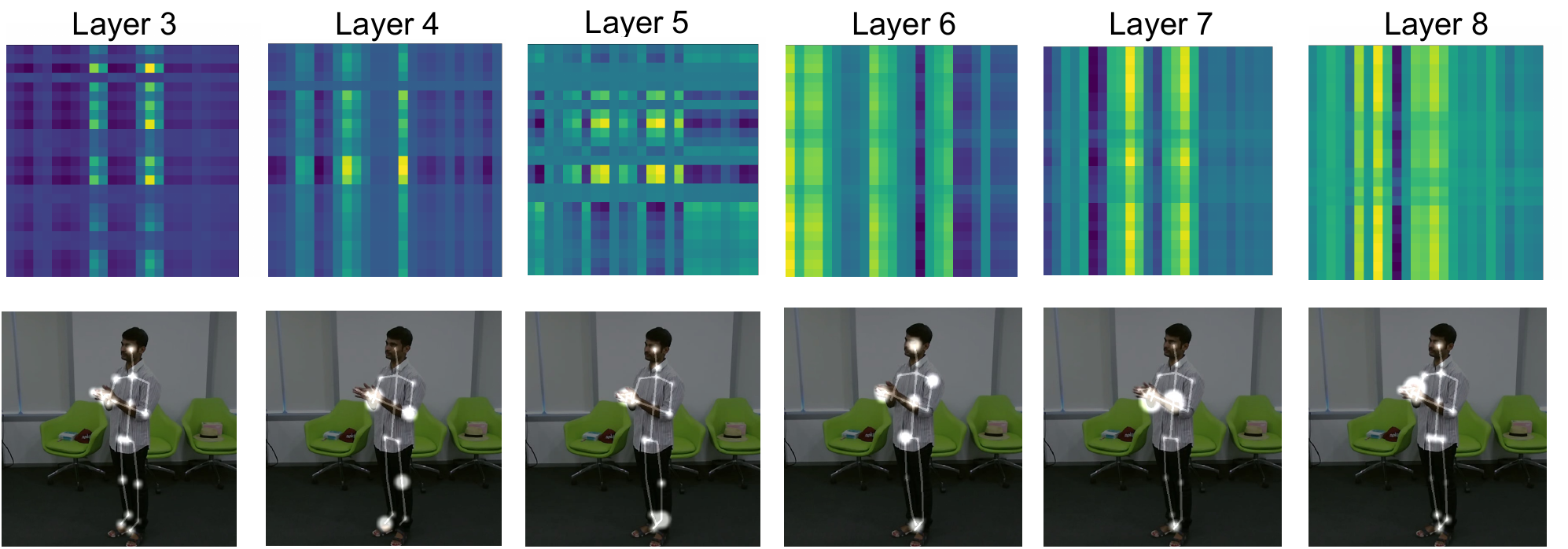}
    \caption{\new{Layer visualization. Each column represent the $k$-th layer, along with the corresponding spatial self-attention heatmap on top, and the skeleton of the subjects performing the action on the bottom.}}
    \label{fig:layers}
\end{figure}

\new{In Figure \ref{fig:heatmaps}, we report some actions and the corresponding Spatial Self-Attention maps. On the top we draw the skeleton of the subjects, where the radius of the circles in correspondence to each joint is proportional its relevance predicted by the self-attention. 
The heatmaps on the bottom represent the attention scores of the last layer; these are $25 \times 25$ matrices, where each row and each column represents a body joint. An element in position $(i,j)$ represents the predicted correlation between joint $i$ and joint $j$ in the same frame. As it can be observed, depending on the action, different parts of the body are activated. In Figure \ref{fig:layers} are shown the same heatmaps \textit{at each layer}. In the first layers, self-attention captures \textit{low-level} correlations between body joints, as highlighted by the sparsity of the activations. While going deeper through the network, the \textit{global} importance of each node emerges instead, as highlighted by the vertical lines corresponding to the most relevant joints. 
}

\section{Conclusions}
In this paper we propose a novel approach that introduces Transformer self-attention in skeleton activity recognition as an alternative to graph convolution. Through extensive experiments on NTU-60, NTU-120 and Kinetics, we demonstrated that our Spatial Self-Attention module (SSA) can replace graph convolution, enabling more flexible and dynamic representations. Similarly, Temporal Self-Attention module (TSA) overcomes the strict locality of standard convolution, enabling the extraction of long-range dependencies across 
the action. \new{Moreover, our final Spatial-Temporal Transformer network (ST-TR)
achieves state-of-the-art performance on all dataset w.r.t. methods using same input joint information and stream setup, and results on-par with state-of-the-art methods when bones information is added.} 
As configurations only involving self-attention modules revealed to be sub-optimal, a possible future work is to search for a unified Transformer architecture able to replace graph convolution in a variety of tasks. 

\bibliographystyle{model2-names}
\bibliography{refs}

\end{document}